\documentclass[conference,a4paper]{IEEEtran}
\IEEEoverridecommandlockouts
\usepackage{algorithmic}
\usepackage{array}
\usepackage[caption=false,font=small,labelfont=rm,textfont=rm]{subfig}
\usepackage[font=small,justification=centering,labelfont=rm,textfont=rm]{caption}
\usepackage{textcomp}
\usepackage{stfloats}
\usepackage{url}
\usepackage{multirow}
\usepackage{amsmath,amssymb}

\usepackage[inline]{enumitem}

\usepackage{verbatim}
\usepackage{graphicx}
\def\BibTeX{{\rm B\kern-.05em{\sc i\kern-.025em b}\kern-.08em
		T\kern-.1667em\lower.7ex\hbox{E}\kern-.125emX}}
\usepackage{balance}
\usepackage[utf8]{inputenc}
\usepackage[ruled,vlined]{algorithm2e}
\usepackage{hyperref}
\hypersetup{
	colorlinks = true, 
	linkcolor = blue,
	urlcolor = blue,
	citecolor = blue
}

\begin{document}
	\title{Personalized Federated Distillation Assisted Vehicle Edge Caching Strategy
	
	\thanks{This work was supported in part by the National Natural Science Foundation of China under Grant No. 61701197, 62531015 and U25A20399; in part by Basic Research Program of Jiangsu under Grant BK20252084; in part by the National Key Research and Development Program of China under Grant No. 2021YFA1000500(4); in part by the Shanghai Kewei under Grant 24DP1500500; and in part by the 111 Project under Grant No. B23008. (Corresponding author: Qiong Wu.)}
	}
	\author{Xun Li\textsuperscript{1}, Qiong Wu\textsuperscript{1,*}, Pingyi Fan\textsuperscript{2}, Kezhi Wang\textsuperscript{3}, Wen Chen\textsuperscript{4}, and Cui Zhang\textsuperscript{5}\\
		\textsuperscript{1}\textit{School of Internet of Things Engineering, Jiangnan University, Wuxi 214122, China} \\
		\textsuperscript{2}\textit{Department of Electronic Engineering, State Key laboratory of Space Network and Communications, Beijing } \\
		\textit{National Research Center for Information Science and Technology, Tsinghua University, Beijing 100084, China} \\
		\textsuperscript{3}\textit{Department of Computer Science, Brunel University, London, Middlesex UB8 3PH, U.K} \\
		\textsuperscript{4}\textit{Department of Electronic Engineering, Shanghai JiaoTong University, Shanghai 200240, China}\\
		\textsuperscript{5}\textit{School of Internet of Things Engineering, Wuxi Institute of Technology, Wuxi, 214121, China}\\ Email: xunli@stu.jiangnan.edu.cn, qiongwu@jiangnan.edu.cn, fpy@tsinghua.edu.cn,\\ Kezhi.Wang@brunel.ac.uk, wenchen@sjtu.edu.cn, zhangcui@wxit.edu.cn}
	
	\maketitle

\begin{abstract}
Vehicle edge caching is a promising technology that can significantly reduce the latency for vehicle users (VUs) to access content by pre-caching user-interested content at edge nodes. It is crucial to accurately predict the content that VUs are interested in without exposing their privacy. Traditional federated learning (FL) can protect user privacy by sharing models rather than raw data. However, the training of FL requires frequent model transmission, which can result in significant communication overhead. Additionally, vehicles may leave the road side unit (RSU) coverage area before training is completed, leading to training failures.
To address these issues, in this paper, we propose a personalized federated distillation assisted vehicle edge caching strategy. The simulation results demonstrate that the proposed vehicle edge caching strategy has good robustness to variations in vehicle speed, significantly reducing communication overhead.


\end{abstract}

\section{Introduction}
\IEEEPARstart{W}{I}{T}{H} the rapid growth of demand for autonomous driving and vehicle networking, vehicles need to obtain map navigation data and multimedia content in real-time to improve driving efficiency and experience \cite{xy}.
Vehicle edge caching technology, by pre-caching content of interest to vehicle users (VUs) in roadside units (RSUs), can significantly reduce content request latency and core network pressure \cite{IEEE2}. However, the cache capacity of RSUs is limited, so it is crucial to accurately predict the content that users are interested in \cite{xun, IEEE1}.

Denoising diffusion probabilistic model (DDPM) has received widespread attention due to its outstanding generative performance \cite{chennan}. 
User data is usually sparse because users only rate a limited number of items. DDPM can learn the distribution of user data and generate high-quality samples to fill in the unrated items.
Applying DDPM to edge caching can achieve more accurate content prediction to a certain extent, enabling vehicles to obtain content that better meets their needs at RSUs. However, the training and deployment of DDPM require substantial computational resources, making them unsuitable for resource-constrained vehicle environments. 
The lightweight U-Net architecture of the DDPM has been proven to effectively reduce its demand for computing resources \cite{grad}. In \cite{xun}, Xun \textit{et al.} adopted a lightweight DDPM (LDPM) and achieved promising performance in edge caching. Therefore, in this paper, we adopt LDPM.

However, model training still requires users’ personal data, which raises privacy concerns.
Federated learning (FL) can protect user privacy by sharing local models rather than raw data \cite{fedavg}. Nevertheless, each round of iteration in FL requires the transmission of full model parameters between vehicles and the server, which incurs significant communication overhead. Federated Distillation is an innovative algorithmic paradigm of FL. It combines knowledge distillation technology with FL, achieving global model training by sharing model output results instead of model parameters while protecting user privacy. Compared with FL based on parameter exchange, its communication overhead is significantly reduced \cite{fedcache}.

Moreover, the high mobility of vehicles also poses significant challenges to edge caching. Vehicles may leave the coverage area of the RSU before the model training is completed, resulting in the failure of model training. Previous methods usually perform vehicle selection before training to ensure that training can be completed \cite{zy}. However, this ignores the distribution of user data for vehicles that did not participate in the training. To the best of our knowledge, no existing studies have adequately addressed this problem. In this paper, we propose a personalized federated distillation assisted vehicle edge caching strategy \footnote{The source code has been released at: https://github.com/qiongwu86/Federated-Distillation-Assisted-Vehicle-Edge-Caching-Scheme-Based-on-Lightweight-DDPM}, 
where each vehicle trains a personalized model locally and only needs to exchange hash-to-index ($HI$) and knowledge-to-index ($KI$) pairs with the RSU. Even if the vehicle moves out of the RSU's coverage area, the training remains unaffected. Moreover, compared with FL based on parameter exchange, the data volume of $HI$ and $KI$ pairs is significantly smaller, which can greatly reduce the communication overhead \cite{FD}. The main innovations of this paper are as follows: 
\begin{itemize}
	\item We propose a personalized federated distillation local model training strategy that enables each vehicle to train a personalized model locally. Specifically, we construct the cosine similarity between the target VU and other VUs to find neighbor users with similar interests, and then pass the knowledge of the neighbor users to the target VU for its local distillation training.
	
	\item A mobile-aware cache content replacement strategy is adopted. Specifically, we jointly determine the weight of a vehicle's preferred content in the RSU based on the vehicle's position and speed factors within the RSU.
	
	\item We propose a hash knowledge ($HK$) module update sharing strategy. RSUs periodically upload the content in their $HK$ modules to the macro base station (MBS). The MBS aggregates the $HK$ cache content of the RSUs within its coverage area, retains only the latest $HK$ cache of each vehicle based on the vehicle's upload time, and then synchronizes all vehicles' latest $HK$ caches to each RSU.
\end{itemize}

This paper is structured as follows: Section \ref{sec2} introduces the system model. Section \ref{sec3} presents the proposed vehicle edge caching strategy. Section \ref{sec4} conducts relevant experiments. Section \ref{sec5} concludes the paper.

\vspace{-1pt} 
\section{System Model}
\label{sec2}
\subsection{System Scenario}
\begin{figure}
	\center
	\includegraphics[scale=0.39]{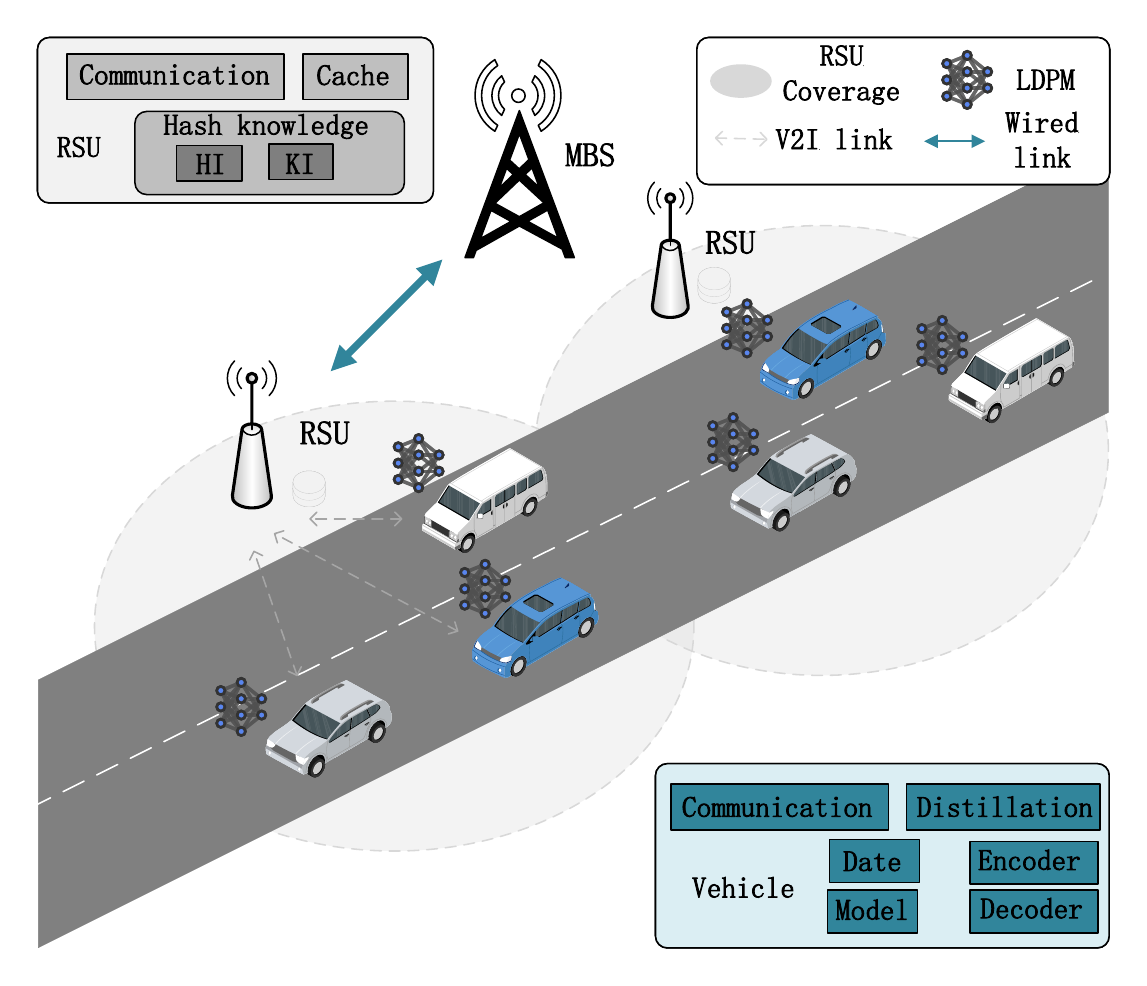}
	\vspace{-0.2cm}
	\caption{System Model.}
	\label{system model}
\end{figure}

We consider a highway system scenario as illustrated in Fig. \ref{system model}, where the vehicle edge computing network consists of three layers: MBS, RSUs, and vehicles. Each RSU $r = 1, 2, \ldots, R$ is connected to the MBS via a reliable wired link. Each vehicle $i = 1, 2, \ldots, I$ is located within the coverage area of the RSU and has a corresponding VU. Each vehicle has its own local data $d_i$, and is equipped with a communication module, a distillation module, and locally pre-trained encoder and decoder networks. The speed of each vehicle follows an independent and identically distributed truncated Gaussian distribution. 
Let $V_i^r$ denote the speed of vehicle $i$ under $r$-th RSU, then the probability density function $f({V_{i}^r})$ of $V_i^r$ is expressed as \cite{2021cache}
\vspace{-3pt} 
\begin{equation}
	f({V_{i}^r}) = \left\{ \begin{aligned}
		\frac{{{2e^{ - \frac{1}{{2{\sigma ^2}}}{{({V_{i}^r} - \mu )}^2}}}}}{{\sqrt {2\pi {\sigma ^2}} (erf(\frac{{{V_{\max }} - \mu }}{{\sigma \sqrt 2 }}) - erf(\frac{{{V_{\min }} - \mu }}{{\sigma \sqrt 2 }}))}}\\
		{V_{min }} \le {V_{i}^r} \le {V_{max }},\\
		0 \qquad \qquad \qquad \qquad \quad otherwise,
	\end{aligned} \right.
	\label{eq1}
\end{equation}
where $V_{max}$ and $V_{min}$ are the maximum and minimum speed thresholds of the vehicle, and $erf(\frac{{{V_i^r} - \mu }}{{\sigma \sqrt 2 }})$ is the Gaussian error function with mean $\mu$ and variance $\sigma^{2}$.

Each RSU is equipped with a communication module, a caching module, and a $HK$ module. The $HK$ module consists of a $HI$ module and a $KI$ module. The $HI$ module is used to store $HI$ pairs, where each $HI$ pair consists of a hash code and a user index. The hash code is encoded from the VU's local data and is used to calculate the cosine similarity between the VU and other VUs to find neighbor users with similar preferences. The $KI$ module is used to store $KI$ pairs, where each $KI$ pair is composed of model output knowledge and a vehicle index.
The caching module has limited caching capacity, capable of caching at most $N$ pieces of content. We assume that the MBS stores all available content. When the content requested by a vehicle is cached in the RSU, the RSU directly delivers the content to the vehicle. Otherwise, the RSU requests the content from the MBS before delivering it to the vehicle, which results in higher content request latency. Our goal is to reduce the latency for VUs to access content by accurately predicting the content of interest to VUs and pre-caching it at the RSUs, thereby maximizing the cache hit percentage of the cached content.
\subsection{Denoising Diffusion Probabilistic Model (DDPM)}
The theoretical foundation of DDPM stems from the entropy increase-inverse process of non-equilibrium thermodynamic systems. DDPM achieves forward diffusion process and reverse diffusion process through parameterized Markov chains.

\subsubsection{Forward Diffusion Process}
The forward process parameterizes the Markov chain using a dynamic scheduling policy $\{\beta_t\}_{t=1}^T$, gradually adding Gaussian noise to the original data, causing the data distribution to gradually approach random noise perturbation, where $T$ is the number of time steps. The sequence $\{\beta_t\}_{t=1}^T$ gradually increases from $\beta_1=10^{-4}$ to $\beta_T=0.02$. Let $\bar{\alpha}_t = \prod_{i=1}^t\alpha_i$ and $\alpha_t = 1-\beta_t$, then we can obtain $q\left(\mathbf{x}_t|\mathbf{x}_0\right)$ and $\mathbf{x}_t$ as follows
\vspace{-2pt}
\begin{equation}
	q\left(\mathbf{x}_t|\mathbf{x}_0\right)=\mathcal{N}\left(\mathbf{x}_t;\sqrt{\overline{\alpha}_t}\mathbf{x}_0,\left(1-\overline{\alpha}_t\right)\mathbf{I}\right),\end{equation}
\vspace{-2pt} 
\begin{equation}
	\mathbf{x}_t=\sqrt{\overline{\alpha}_t}\mathbf{x}_0+\sqrt{1-\overline{\alpha}_t}\boldsymbol{\epsilon}_t, \boldsymbol{\epsilon}_t\sim\mathcal{N}\left(0,\mathbf{I}\right).\end{equation}
Here, $x_t$ represents the data at time step $t$, $\mathcal{N}(\cdot)$ represents a Gaussian distribution and $\mathbf{I}$ is the identity matrix.
\subsubsection{Reverse Diffusion Process}
The reverse diffusion process gradually denoises the noise to restore the original data distribution. We train a neural network $\boldsymbol{\mu}_\theta$ to predict the noise at each step. Each step of the reverse diffusion process $p_\theta\left(\mathbf{x}_{t-1}|\mathbf{x}_t\right)$ can be defined as
\begin{equation}
	p_\theta\left(\mathbf{x}_{t-1}|\mathbf{x}_t\right)=\mathcal{N}\Big(\mathbf{x}_{t-1};\boldsymbol{\mu}_\theta\left(\mathbf{x}_t,t\right),\frac{1-\bar{\alpha}_{t-1}}{1-\bar{\alpha}_t}\beta_t\Big),
\end{equation}
\begin{equation}
	\boldsymbol{\mu}_\theta\left(\mathbf{x}_t,t\right)=\frac{1}{\sqrt{\alpha_t}}\Big(\mathbf{x}_t-\frac{\beta_t}{\sqrt{1-\overline{\alpha}_t}}\boldsymbol{\epsilon}_\theta\left(\mathbf{x}_t,t\right)\Big).
\end{equation}

In \cite{ddpm}, Ho \textit{et al.} proposed a simplified objective function for optimization, expressed as
\begin{equation}
	\mathcal{L}_{t-1}^{simple}=\mathbb{E}_{t,\mathbf{x}_0,\boldsymbol{\epsilon}\sim\mathcal{N}(0,\mathbf{I})}\Big[\Big\|\boldsymbol{\epsilon}-\boldsymbol{\epsilon}_\theta\Big(\sqrt{\overline{\alpha}_t}\mathbf{x}_0+\sqrt{1-\overline{\alpha}_t}\boldsymbol{\epsilon},t\Big)\Big\|^2\Big].
	\label{eq7}
\end{equation}

\section{Vehicle Edge Caching Scheme}  
\label{sec3}
In this section, we introduce the proposed personalized federated distillation local model training strategy, which enables training a personalized local model for each vehicle. By constructing the cosine similarity between the target VU and other VUs, we identify neighbor users with preferences similar to the target VU, and then transfer the knowledge of these neighbor users to the target VU for its local distillation training. During the local training process, only small-data-volume $HI$ pairs and $KI$ pairs need to be exchanged between vehicles and RSUs, leading to extremely short interaction time between them. Moreover, since the model training is conducted locally, the local training process can still proceed normally even if a vehicle moves out of the RSU coverage area. Therefore, our strategy significantly reduces the requirements for the connection time between vehicles and RSUs during model training.
This scheme includes the execution process between vehicles and RSUs, and the execution process between RSUs and MBS. We will describe it in the following parts.
\subsection{Execution between Vehicles and RSUs}
\begin{figure}
	\center
	\includegraphics[scale=0.34]{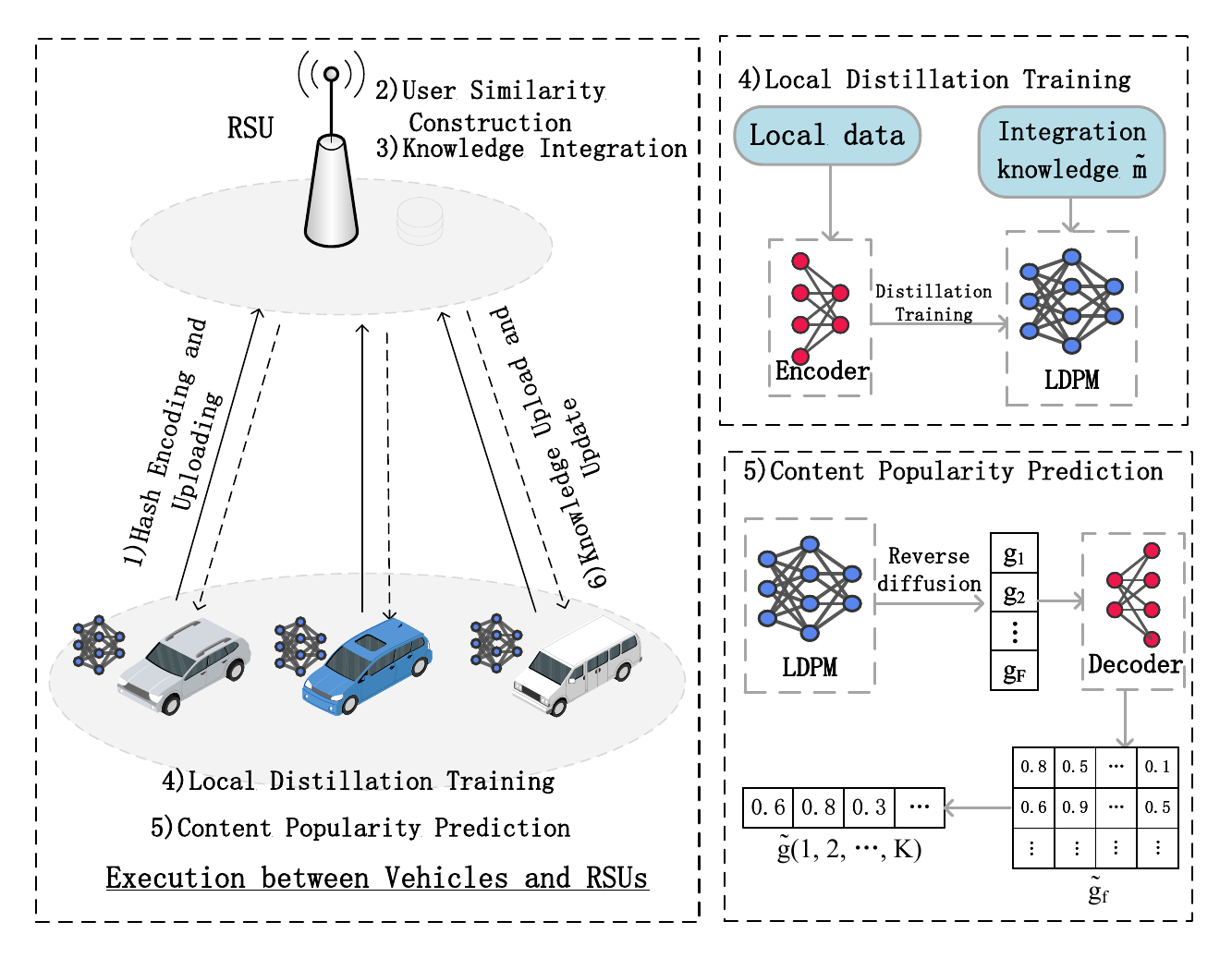}
	\vspace{-0.2cm}
	\caption{Execution between Vehicles and RSUs.}
	\label{system model comprehensive2}
\end{figure}

Each vehicle $i$ will upload its content recommendation list ${L}_{i,r}$ when entering $r$-th RSU, where ${L}_{i,r}$ denotes the content recommendation list of vehicle $i$ in the $r$-th RSU. The content recommendation list is generated by the vehicle using local models. The execution between the RSUs and the vehicles will be continuously repeated, as shown in Fig. \ref{system model comprehensive2}, which includes the following steps:

\subsubsection{Hash Encoding and Uploading}
Each vehicle uploads the $HI$ pair to the RSU. The $HI$ pair consists of a hash code $h_i^r$ and a vehicle index $i$, where $h_i^r$ represents the hash code of vehicle $i$ in the $r$-th RSU. The hash code $h_i^r$ is generated by the vehicle locally using a pre-trained encoder \( E_i(\cdot) \) to encode their local data $d_i$, i.e.
\begin{equation}
	h_i^r = E_i(d_i).
	\label{eq3}
\end{equation}
The pre-encoder is designed as a deep neural network, and its output dimension is much smaller than that of the original data, thus sharing hash code with RSU ensures privacy and also having low communication overhead. Pre-trained encoder and pre-trained decoder are trained using publicly available datasets and then fine-tuned with local data \cite{laten}.
\subsubsection{User Similarity Construction}
After the RSU receives the $HI$ pair uploaded by vehicle $i$, it immediately saves or updates the corresponding content in the $HI$ module,
\begin{equation}
	HI(i,r) \leftarrow (h_i^r, i).
	\label{eq4}
\end{equation}
where $HI(i,r)$ refers to the $HI$ pair of VU $i$ under the $r$-th RSU.

Directly using the knowledge of all VUs for a target VU's local distillation training would make the model learn the interests of all VUs rather than the target VU. Therefore, we propose computing the cosine similarity between VUs to find neighboring users with similar interest preferences and using those neighbors' knowledge for the target VU's distillation training.
Since the data of VUs with similar preferences are mapped to hash code vectors with similar directions, calculating the cosine similarity of $HI$ pairs between VUs can find neighbor users with similar preferences. For any VU $i$ and $j$, their cosine similarity can be calculated as \cite{HSWN}
\begin{equation}
\begin{split}
	\operatorname{sim}_{i}^{j} =  \cos \left (HI(i,r), HI(j,r)\right)\\
	= \frac{\boldsymbol{HI}(i,r) \cdot \boldsymbol{HI}(j,r)^T}{\left\|\boldsymbol{HI}(i,r)\right\|_{2} \times\left\|\boldsymbol{HI}(j,r)\right\|_{2}},
	\label{eq5}
\end{split}
\end{equation}
where $\left\|\boldsymbol{HI}(i,r)\right\|_{2}$ and $\left\|\boldsymbol{HI}(j,r)\right\|_{2}$ are the 2-norm of $\boldsymbol{HI}(i,r)$ and $\boldsymbol{HI}(j,r)$, respectively. For any VU $i$, select the top $C$ VUs with the highest cosine similarity to it as its neighbor users, and these neighbor users share similar preferences with VU $i$.
\subsubsection{Knowledge Integration}
The RSU extracts knowledge $(m_i^r)_{c}$ from the $KC$ for the $C$ neighboring users of VU $i$, where $(m_i^r)_{c}$ is the knowledge of the $c$-th similar neighbor user of VU $i$ in $r$-th RSU. 
Then, this knowledge is averaged to obtain aggregated knowledge $\tilde{m}_i^r$ that reflects the overall preferences of neighboring users, subsequently passing it to the corresponding VU,
\begin{equation}
	\vspace{-2pt} 
	\tilde{m}_i^r = \frac{1}{C} \sum_{c=1}^{C} (m_i^r)_c.
	\label{eq7}
\end{equation}

\subsubsection{Local Distillation Training}
In this step, each vehicle $i$ uses local data and acquired integrated knowledge $\tilde{m}_i^r$ to conduct local distillation training. Since latent-space training with LDPM can significantly reduce the required computational resources and accelerate training efficiency \cite{laten}. We map the original data $d_i$ to a low-dimensional latent space $\hat{d}_i=E_i(d_i)$ before training. The local LDPM is then trained in this low-dimensional latent space. For each data $\hat{z}$ in $\hat{d}_i$, the optimization function for local distillation training is calculated as \cite{fedcache}
\begin{equation}
	\begin{split}
		\arg\min_{\omega_i} J^i(\omega_i)
		&= \arg\min_{\omega_i} 
		\sum_{\hat{z} \in \hat{d}_i}\Bigl[
		\mathcal{L}\bigl( \omega_i; \hat{z} \bigr) \\
		\;+\;
		&\lambda\,\mathrm{KL}\bigl(\,o\bigl(g_i^r ;\delta\bigr)\;\big\|\;o\bigl(\tilde{m}_i^r ;\delta\bigr)
		\bigr)
		\Bigr].
	\end{split}
\label{eq8}
\end{equation}
The first part is the original training loss of the local model for its own samples, where $\omega_i$ is the local LDPM parameter; the second part is the KL divergence term that incorporates integrated knowledge into training, where $\delta$ is the temperature parameter of the softmax function $o\bigl(\cdot\bigr)$, $\lambda$ is the weight factor, and $g_i^r$ is the output of the local model.

\subsubsection{Content Popularity Prediction}
After completing the local distillation training, each VU $i$ uses the local LDPM to perform the reverse diffusion process, generating $F$ sample data points $g_{i,f}^r$, where $F$ is the number of sample data points and $f = 1, 2, \ldots, F$. These sample data points are mapped to the original data dimension via the pre-trained decoder to obtain the reconstructed sample data $\tilde{g}_{i,f}^r = D(g_{i,f}^r)$, where \( D(\cdot) \) represents the decoder network. These reconstructed data are used for content popularity prediction. Compared to the original data, these reconstructed data better reflect VUs' preferences for content because they contain fewer zero values \cite{zy}.
Assuming there are a total of $K$ content items, the dimension of $\tilde{g}_{i,f}^r$ is $K$ and can be expressed as $\tilde{g}_{i,f}^r(1,2,...,K)$. All reconstructed data can be added by dimension to obtain the score $\tilde{g}_i^r(1,2,...,K)$ of all contents,
\begin{equation}
	\vspace{-1pt} 
	\tilde{g}_i^r(1,2,...,K)=\frac{1}{F}\sum_{f=1}^{F}\tilde{g}_{i,f}^r(1,2,...,K).
	\label{eq9}
\end{equation}
The score $\tilde{g}_i^r(1,2,...,K)$ reflects the preferences of VU $i$. The higher the score, the more interested the VU is. Then, select the top $M$ items as the VU's new content recommendation list ${L}_{i,r}$.

\subsubsection{Knowledge Upload and Update}
The vehicle uploads $KI$ pair and the new content recommendation list to the RSU. The $KI$ pair is constructed using the vehicle index $i$ and the model output knowledge $g_{i,f}^r$,
\begin{equation}
\vspace{-1pt} 
KI(i,r) \leftarrow (g_{i,f}^r, i).
\label{eq10}
\end{equation}
Subsequently, the RSU updates the $KI$ module to ensure that the latest knowledge can be retrieved for the next request.
Since the dimension of the knowledge is much lower than that of the VU's original data, sharing the knowledge will not leak the VU's privacy and having low communication overhead.

\subsection{Execution between RSUs and MBS}
When vehicles within the coverage of an RSU uploads a content recommendation list, the RSU will update its cache content. It is unfair for the RSU to treat the content recommendation lists of different vehicles equally during cache replacement, since vehicles that stay longer in the RSU are more likely to request more content. Therefore, we adopt a mobility-aware cache content replacement strategy. Additionally, to prevent the information stored in $HK$ module from becoming outdated after vehicles leave the RSU's coverage, we propose a periodic update strategy for the $HK$ module. The execution between the RSU and MBS are as follows:
\subsubsection{Mobility-Aware Cache Content Replacement Strategy}
The RSU updates the cached content based on the content recommendation list uploaded by vehicles within its coverage. In addition, the position of the vehicle in the RSU and the vehicle speed factors should also be considered, as vehicles that stay in the RSU for a longer time are more likely to request more content, and their content recommendation list should be given greater weight. For each content $k$, its score $u_r(k)$ can be represented as
\begin{equation}
	u_r(k)=\sum_{i\in \mathcal{U}_r} \eta\frac{B-P_i}{V_i^r} \mathbf{1}(k\in {L}_{i,r}), 
	\label{eq11}
\end{equation}
where $\mathcal{U}_r$ is the set of vehicles currently served by $r$-th RSU, $P_i$ is the distance from vehicle $i$ to the entrance of RSU, $\eta$ is the location weight factor, $B$ is the coverage range of RSU and $1\bigl(\cdot\bigr)$ is an indicator function. The RSU caches the top $N$ contents with the highest scores based on the scores of all contents.
\subsubsection{KC Update Sharing Strategy}
The RSU periodically uploads the content of its $HK$ to the MBS. The MBS aggregates the $HK$ of RSUs within its coverage area, retaining only the latest $HK$ for each vehicle $i$ based on their upload time. It then passes the updated $HK_{lat}$ to the RSUs within its coverage area. The $HK$ update process is as follows
\begin{equation}
	r_i^* = \arg\max_{r\in\{1,\dots,R\}}\,t_{i,r},
	\label{eq12}
\end{equation}
\begin{equation}
	HK_{{lat}}
	= \bigcup_{i=1}^I
	\bigl\{\,
	KI\bigl(i,\,r_i^*\bigr)\,,\;
	HI\bigl(i,\,r_i^*\bigr)
	\bigr\},
	\label{eq13}
\end{equation}
where $t_{i,r}$ represents the time at which vehicle $i$ uploads $HI$ pair to the $r$-th RSU.

\section{Simulation}
\label{sec4}
In this section, we introduce the experiments. The dataset is the widely-used MovieLens 1M, which includes 1,002,099 rating records from 6,040 users on 3,952 movies. These ratings range from 0 to 5. Some parameters of the experiments are shown in Table~\ref{tab1}. Unless otherwise specified, we set the cache capacity as 500 and the vehicle speed as 25 m/s.
Our overall U-Net architecture is similar to \cite{Unet}. The main differences are the replacement of 2D convolutions with 1D convolutions to adapt to the user’s interaction data structure, and the use of one-fourth of the number of channels and three feature map resolutions to reduce the model size. 
Additionally, to evaluate the performance of the schemes, we introduce cache hit percentage and request content latency. Cache hit percentage indicates the probability of directly accessing the requested content from nearby RSU. If the requested content has been cached in advance, it is considered a successful cache; otherwise, it is a failure. The cache hit percentage can be expressed as \cite{2025cache}
\begin{equation}
	\text{cache hit percentage} = \frac{C_s}{C_s + C_f} \times 100\%,
\end{equation}
where $C_s$ and $C_f$ represent the number of successful cache and failed cache, respectively. The request content delay represents the average delay of all VUs getting content.
\begin{table*}[ht]
	\vspace{8pt}
	\caption{Cache hit percentage and communication overhead under different schemes.}
	\label{tab2}
	\footnotesize
	\centering
	\begin{tabular}{|c|cccccccc|c|}
		\hline
		\multirow{2}{*}{Scheme}  &  \multicolumn{8}{c|}{Cache capacity} &  \multirow{2}{*}{ Overhead (MB)}   \\ 
		\cline{2-9}
		& 150     & 200     & 250     & 300     & 350     & 400     & 450     & 500 &\\  
		\hline
		\textbf{Ours} & \textbf{23.95\%} & \textbf{29.13\%} & \textbf{33.82\%} & \textbf{38.19\%} 
		& \textbf{42.34\%} & \textbf{45.81\%} & \textbf{49.15\%} & \textbf{52.14\%} & \textbf{1.78} \\
		
		Oracle                 & 24.54\%  & 29.91\%  & 34.71\%  & 39.03\%  & 42.97\%  & 46.57\%  & 49.93\% & 53.05\%   & -\\
		
		CAFR                   & 23.93\% & 29.04\% & 33.73\% & 37.96\% & 41.97\% & 45.44\% & 48.59\% & 51.55\%  & 12.72 \\
		
		AsyFed    &  22.68\% & 27.44\% & 32.07\% & 36.19\% & 40.20\% & 43.59\% & 46.97\% & 49.64\%   & 123.18 \\
		
		FedAvg    &  21.49\% & 26.13\% & 30.91\% & 35.06\% & 38.73\% & 42.14\% & 45.47\% & 48.31\% &1849.80  \\
		
		N-$\tau$-greedy   & 20.27\% & 24.94\% & 28.99\% & 32.73\% & 36.19\% & 38.96\% & 42.45\% & 45.07\%   &-\\
		
		\hline
	\end{tabular}
\end{table*}
\begin{table}
	\caption{Values of the parameters in the experiments.}
	\label{tab1}
	\footnotesize
	\centering
	\begin{tabular}{|p{1.3cm}<{\centering}|p{1.2cm}<{\centering}|p{1.3cm}<{\centering}|p{1.2cm}<{\centering}|}
		\hline
		\textbf{Parameter} &\textbf{Value} &\textbf{Parameter} &\textbf{Value}\\
		\hline
		$T$ & $50$ & $\delta$ &$2$\\
		\hline
		$\lambda$ &$1$ & $K$  & $3952$\\
		\hline
		$F$ & $500$ & $\eta$ & $0.1$ \\
		\hline
		$C$ & $5$ & $I$ & $20$ \\
		\hline
		
	\end{tabular}
\end{table}
\begin{figure}[htbp]
	\centering
	\includegraphics[scale=0.41]{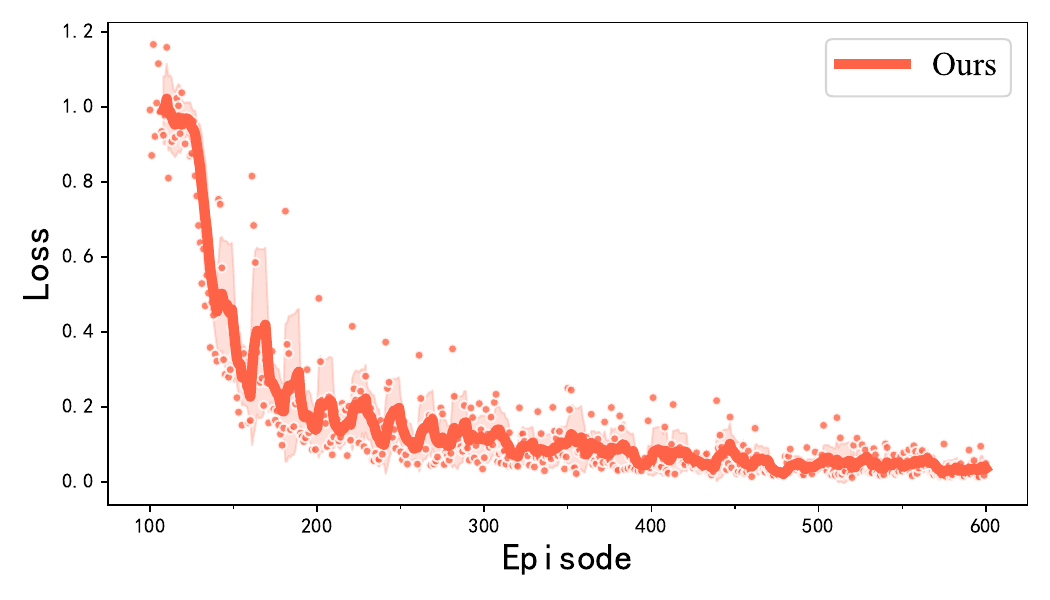}
	\caption{Loss versus episodes.}
	\label{fig1}
\end{figure}
We compare our scheme with other schemes, such as:
\begin{itemize}
	\item Oracle: This scheme possesses full prior knowledge of VUs' future requests, defining the theoretical maximum achievable cache hit percentage.
	\item CAFR \cite{zy}: This scheme proposed the combination of asynchronous FL and autoencoder for edge caching, combined with a mobility aware mechanism, provides a feasible solution for vehicle edge caching.
	\item N-$\tau$-greedy: This scheme selects the contents with the $N$ largest request counts based on probability $1-\tau$ and selects $N$ contents randomly based on probability $\tau$ to cache. In the experiment, $\tau = 0.2$.
	\item FedAvg \cite{fedavg}: This scheme adopts traditional FL algorithm training model instead of our proposed personalized federated distillation strategy.
	\item AsyFed \cite{fedasy}: This scheme adopts the asynchronous federated training model instead of our proposed personalized federated distillation strategy.
	\vspace{-8pt} 
\end{itemize}
\begin{figure}[htbp]
	\centering
	\includegraphics[scale=0.41]{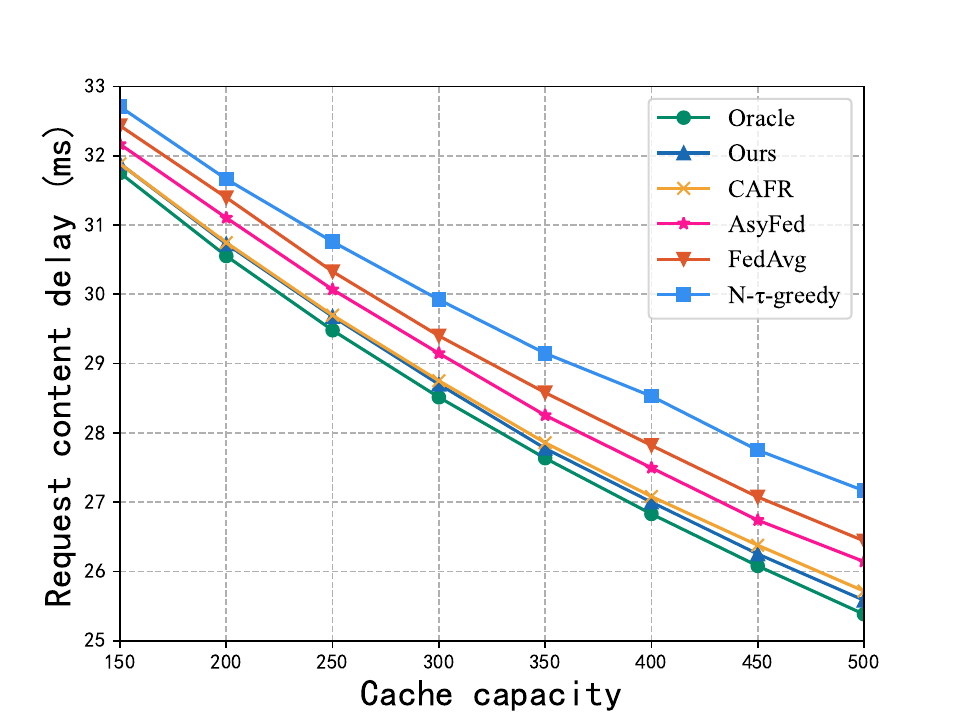}
	\caption{Request content delay versus cache capacity.}
	\label{fig2}
\end{figure} 
Fig. \ref{fig1} shows the convergence curve of model training, where it can be observed that the model loss gradually decreases with the increase in episodes, and the model converges at approximately 400 episodes.

As can be seen from Table~\ref{tab2}, with the cache capacity increases, the cache hit percentage of various schemes gradually rise. This is because a higher cache capacity allows the RSU to cache more content, making it more likely for VUs to request the desired content from the RSU. Oracle achieves the best performance because it knows the VUs' future requests, representing the theoretical optimal value. It can be seen that our scheme achieves the best performance except for Oracle, and this is because our scheme trains a personalized local model for each vehicle, which can accurately predict the content each VU is interested in. AsyFed performs worse than CAFR, as it is difficult for AsyFed to complete training before high-speed moving vehicles leave the RSU. FedAvg is more severely affected by vehicle speed, thus its performance is inferior to that of AsyFed. N-$\tau$-greedy algorithm exhibits poorer performance because they are non-learning-based algorithms. 
Furthermore, it can be seen that the communication overhead of our scheme is much lower than that of other schemes. Compared with the traditional FedAvg and AsyFed schemes, the communication overhead is greatly reduced. This is because our scheme only transmits a small amount of $HI$ pairs and $KI$ pairs instead of the full set of model parameters.
\begin{figure}[htbp]
	\centering
	\includegraphics[scale=0.41]{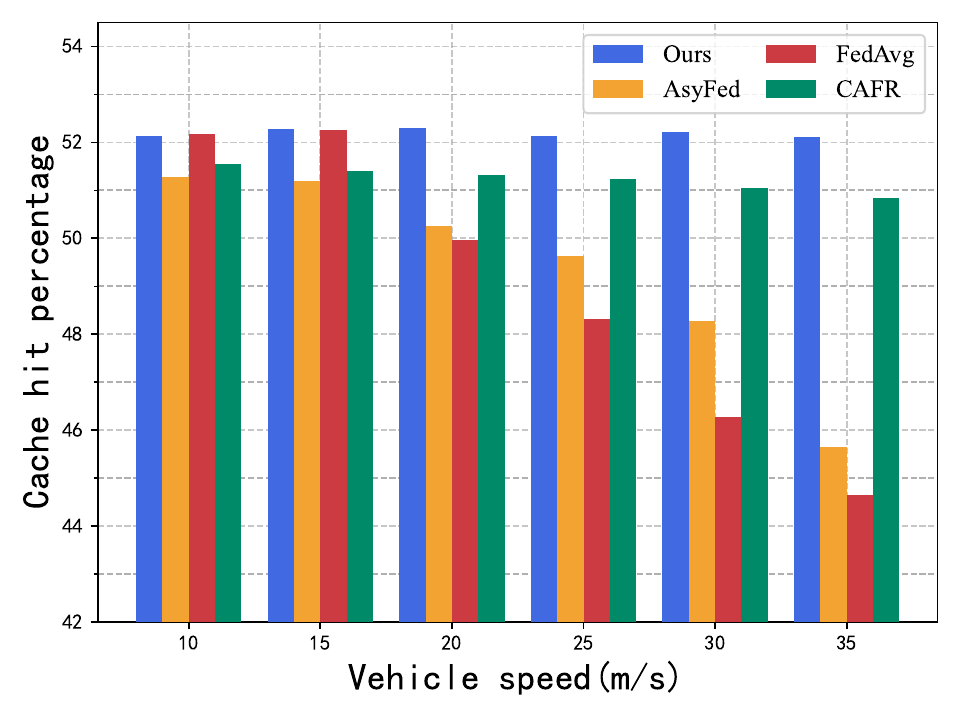}
	\caption{Cache hit percentage versus vehicle speed.}
	\label{fig3}
\end{figure}

Fig. \ref{fig2} shows the changes in request content latency for various schemes under different cache capacities. It can be seen that as the cache capacity increases, the request content latency for all schemes gradually decreases. This is because a larger cache capacity makes it more likely for vehicles to obtain the requested content from nearby RSU, thereby reducing the request content latency. Additionally, it can be observed that our scheme achieves the lowest request content latency, second only to Oracle. This is because our scheme has a higher cache hit percentage, making it more likely for VUs to request the content they are interested in, which reduces the request content latency.


Fig. \ref{fig3} shows the maximum cache hit percentage of various schemes under different vehicle speeds. It can be seen that our scheme is hardly affected by vehicle speed. This is because our scheme trains models locally and does not require frequent interaction with RSUs; even when entering a new RSU, it can still immediately upload the content recommendation list that matches VU preferences. In addition, we also take into account the mobility characteristics of vehicles to update the cache content.
The cache hit percentage of CAFR gradually decreases as the vehicle speed increases. This is because with the rise of vehicle speed, the number of vehicles meeting the requirements of the vehicle selection phase gradually reduces, resulting in fewer vehicles participating in training. As vehicle speed increases, the cache hit percentage of FedAvg and AsyFed gradually decreases. This is because vehicles leave the RSU too quickly, making it impossible to complete the training.

\section{Conclusion}
\label{sec5}
In this paper, we proposed a personalized federated distillation assisted vehicle edge caching strategy, which enables training a personalized local model for each vehicle. By constructing the cosine similarity between the target VU and other VUs, we identify neighbor users with preferences similar to the target VU, and then transfer the knowledge of these neighbor users to the target VU for its local distillation training. Finally, experiments were conducted to verify the effectiveness of the proposed strategy.


	\bibliographystyle{IEEEtran}  

	\bibliography{IEEEabrv,refer.bib}

\vfill
	
\end{document}